\documentclass[11pt,a4paper]{article}
\pdfoutput=1

\usepackage[dvipsnames]{xcolor}
\usepackage{hyperref}
\usepackage[utf8]{inputenc}
\usepackage{acl2017}
\usepackage{times}
\usepackage{latexsym}
\usepackage{amsmath}
\usepackage{amsfonts}
\usepackage{amssymb}
\usepackage{comment}
\usepackage{cleveref}
\usepackage{bm}
\usepackage{graphicx}
\usepackage{booktabs}
\usepackage{url}
\usepackage{enumitem}
\usepackage{siunitx}

\setlist{nosep} 

\newcommand\given[1][]{\:#1\vert\:}

\newcommand{\redbf}[1]{\textbf{\textcolor{Red}{#1}}}

\newcommand{\green}[1]{\textcolor{OliveGreen}{#1}}
\newcommand{\greenbf}[1]{\textbf{\textcolor{OliveGreen}{#1}}}

\crefformat{section}{§#2#1#3}

\aclfinalcopy 
\def\aclpaperid{***} 

\title{Multilingual Multi-modal Embeddings for Natural Language Processing}

\author{Iacer Calixto\\
	    ADAPT Centre\\
	    Dublin City University\\
	    Glasnevin, Dublin 9\\
	    {\small\tt iacer.calixto@adaptcentre.ie}\\
	  \And
	Qun Liu\\
	ADAPT Centre\\
  	Dublin City University\\
  	Glasnevin, Dublin 9\\
  \And
	Nick Campbell\\
	ADAPT Centre\\
  	Trinity College Dublin\\
  	College Green, Dublin 2\\
  }

\date{}

\begin{document}

\maketitle

\begin{abstract}
 We propose a novel discriminative model that learns embeddings from multilingual and multi-modal data, meaning that
 our model can take advantage of images and descriptions in multiple languages to improve embedding quality.
 To that end, we introduce a modification of a pairwise contrastive estimation optimisation function as our training objective.
 We evaluate our embeddings on an image--sentence ranking (ISR), a semantic textual similarity (STS), and a neural machine translation (NMT) task.
 We find that the additional multilingual signals lead to improvements on both the ISR and STS tasks, and the discriminative cost can also be used in re-ranking $n$-best lists produced by NMT models, yielding strong improvements.
\end{abstract}

\setlength{\belowdisplayskip}{1.0pt} \setlength{\belowdisplayshortskip}{1.0pt}
\setlength{\abovedisplayskip}{1.0pt} \setlength{\abovedisplayshortskip}{1.0pt}

\section{Introduction}
\label{sec:intro}

In this work, we expand on the idea of training multi-modal embeddings
\cite{Kirosetal2014b,Socheretal2014} and introduce a model that can be trained not only on images and their monolingual descriptions but also on additional multilingual image descriptions when these are available.
We believe that having multiple descriptions of one image, regardless of its language, is likely to increase the coverage and variability of ideas described in the image, which may lead to a better generalisation of the depicted scene semantics.
Moreover, a similar description expressed in different languages may differ in subtle but meaningful ways.

To that end, we introduce a novel training objective function that uses noise--contrastive estimation adapted to the case of three or more input sources, i.e. an image and multilingual sentences (\cref{sec:multilingual-and-multimodal-embeddings}).
Our objective function links images and multiple sentences in an arbitrary number of languages, and we validate our idea in experiments where we use the Multi30k data set (\cref{sec:data}).

We evaluate our embeddings in three different tasks (\cref{sec:experiments}): an image--sentence ranking (ISR) task, in both directions, where we find that it improves ISR to a large extent, i.e. the median ranks for English are improved from 8 to 5 and for German from 11 to 6, although the impact on ranking sentences given images is less conclusive;
two sentence textual similarity (STS) tasks, finding consistent improvements over a comparable monolingual baseline;
and also in neural machine translation (NMT), where we use our model to re-rank $n$-best lists generated by an NMT baseline and report strong and consistent improvements.
Our main contributions are:
\begin{itemize}
  \item we introduce a novel ranking loss objective function to train a discriminative model that utilises not only \emph{multi-modal} but also \emph{multilingual} data;
  \item we compare our proposed multilingual and multi-modal embeddings (MLMME) to embeddings trained on only one language on three different tasks (ISR, STS and NMT), and find that it leads to improvements on both ISR and STS tasks, and that it can be efficiently used to re-rank NMT $n$-best lists.
\end{itemize}

\section{Multilingual and multi-modal embeddings}\label{sec:multilingual-and-multimodal-embeddings}

Our model has two main components: one \emph{textual} and one \emph{visual}.
In the textual component,
we have $K$ different languages $L_k$, $k \in K$, and for each language
we use
a recurrent neural network (RNN) with 
gated recurrent units (GRU)~\cite{Choetal2014}
as a sentence encoder.
Let $S^{k} = \{w^k_1, \dots, w^k_{N_K}\}$ denote sentences composed of word indices in a language $L_k$, and $X^k = (\bm{x}^k_1, \bm{x}^k_2, \cdots, \bm{x}^k_{N_k})$ the corresponding word embeddings for these sentences, where $N_k$ is the sentence length.
An RNN $\Phi^k_{\text{enc}}$ reads $X^k$ word by word, from left to right, and generates a sequence of annotation vectors ${(\bm{h}^k_1, \bm{h}^k_2, \cdots, \bm{h}^k_{N_k})}$ for each embedding $\bm{x}^k_i$, ${i \in [1,N_k]}$.
For any given input sentence,
we use the corresponding encoder RNN's last annotation vector $\bm{h}^k_{N_k}$ for that language $L_k$ as the sentence representation, henceforth $\bm{v}^k$.


In our visual component we use publicly available pre-trained models for image feature extraction.
~\newcite{SimonyanZisserman2014} trained deep convolutional neural network (CNN) models for classifying images into one out of $1000$ ImageNet classes~\cite{Russakovskyetal2014}.
We use their 19-layer VGG network (VGG$19$) to extract image feature vectors for all images in our dataset:
we feed an image to the pre-trained network and use the activations of the penultimate fully connected layer FC$7$, network configuration E, as our image feature vector~\cite{SimonyanZisserman2014}.


Each training example consists of a tuple \emph{(i)} sentences $S^{k}$ in $L_k$, $\forall k \in K$, and \emph{(ii)} the associated image these sentences describe.
Given a training instance,
we retrieve the embeddings $X^k = \{\bm{x}^k_1, \dots, \bm{x}^k_{N_K}\}$ for each sentence $S^{k}$ using one separate word embedding matrix for each language $k$.
A sentence embedding representation $\bm{v}^k$ is then obtained by applying the encoder $\Phi^k_{\text{enc}}$ onto each embedding $\bm{x}^k_{1:N_K}$ and using the last annotation vector $\bm{h}^k_{N_K}$ of each RNN, after it has consumed the last token in each sentence.
Note that our encoder RNNs for different languages share no parameters.
An image feature vector \mbox{$\bm{q} \in \mathbb{R}^{4096}$} is extracted using a pre-trained CNN so that \mbox{$\bm{d} = W_I \cdot \bm{q}$} is an image embedding and $W_I$ is an image transformation matrix trained with the model.
Also, image embeddings $\bm{d}$ and sentence embeddings $\bm{v}^k$, $\forall k \in K$
are all normalised to unit norm and have the same dimensionality.
Finally, \mbox{$s_i(\bm{d},\bm{v}^k) = \bm{d}^\top \cdot \bm{v}^k$, $\forall k \in K$} is a function that computes the similarity between images and sentences in all languages, and \mbox{$s_s(\bm{v}^k,\bm{v}^l) = (\bm{v}^k)^\top \cdot \bm{v}^l$, $\forall k \in K, \forall l \in K, k \neq l$}, computes the similarity between sentences in two different languages.

We now describe a variation of a \textit{pairwise ranking loss} function used to train our model.
Our model takes into consideration not only the relation between sentences in a given language and images---computed by the $s_i(\cdot,\cdot)$ function---, but also sentences in different languages in relation to each other, computed by $s_s(\cdot,\cdot)$.
Our sentence--image, \emph{multi-modal} similarity is given in~(\ref{eq:text_image_component}):

{\small
\begin{align}\label{eq:text_image_component}
    S_I(\bm{v}^k, \bm{d}) = & \sum_d \sum_r \max{ \{0, \alpha - s_i(\bm{d},\bm{v}^k) + s_i(\bm{d},\bm{v}^k_r)\} } + \notag\\
    & \sum_{v^k} \sum_r \max{ \{0, \alpha - s_i(\bm{v}^k,\bm{d}) + s_i(\bm{v}^k,\bm{d}_r)\} }, \notag\\
    & \forall k \in K,
\end{align}}

\noindent
where $\bm{v}^k_r$ (subscript \textit{r} for \textit{random}) is a contrastive or non-descriptive sentence embedding in language $L_k$ for image embedding $\bm{d}$ and vice-versa, and $\alpha$ is a model parameter, i.e. the \textit{margin}.
$S_I(\bm{v}^k, \bm{d})$ scores sentences in every language $L_k$, $\forall k \in K$ against all images $\bm{d}$.
Our sentence--sentence, \emph{multilingual} similarity is given in~(\ref{eq:text_text_component}):

{\small
\begin{align}
\label{eq:text_text_component}
    S_S(\bm{v}^k) = & \sum_{v^k} \sum_r \max{ \{0, \alpha - s_s(\bm{v}^k,\bm{v}^l) + s_s(\bm{v}^k,\bm{v}^l_r)\} } + \notag\\
    & \sum_{v^l} \sum_r \max{ \{0, \alpha - s_s(\bm{v}^l,\bm{v}^k) + s_s(\bm{v}^l,\bm{v}^k_r)\} }, \notag\\
    & \forall k \in K,  \forall l \in K, l \neq k,
\end{align}
}

\noindent
where $\bm{v}^k_r$ is a contrastive or non-descriptive sentence embedding in language $L_k$ for sentence $\bm{v}^l$ in language $L_l$ and vice-versa.
$S_S(\bm{v}^k)$ scores sentences in one language against sentences in all other languages.
In both $S_I$ and $S_S$, contrastive terms are chosen randomly from the training set and resampled at every epoch.

Finally, our optimisation function
in Equation~(\ref{eq:our_pairwise_loss})
minimises the linearly weighted combination of $S_I$ and $S_S$:

{\small
\begin{align}
\label{eq:our_pairwise_loss}
  & \min_{\theta_k, W_I} \beta S_I + (1 - \beta) S_S, & &            & \forall k \in K, & \\
 & & & & 0 \geq \beta \geq 1, & \notag
\end{align}
}

\noindent
where $\theta_k$ includes all the encoder RNNs parameters for language $L_k$, and $W_I$ is an image transformation matrix.
$\beta$ is a model hyperparameter that controls how much influence a particular similarity (\textit{multi-modal} versus \textit{multilingual}) has in the overall cost.
The two extreme scenarios are $\beta$ = $0$, in which case only the multilingual similarity is used, and $\beta$ = $1$, in which case only the multi-modal similarity is used.
If the number of languages $K$ = $1$ and $\beta$ = $1$, our model computes the Visual Semantic Embedding (VSE) of~\newcite{Kirosetal2014b}.

\begin{table*}[t!]
  \centering
  \resizebox{\linewidth}{!} {
  \begin{tabular}{ c|c|c|c|c|c|c|c|c|c|c|c }
    
    
        
    
    

    
    \multicolumn{7}{c|}{English} &
    \multicolumn{5}{c}{German} \\
    \hline
    
    & \multicolumn{2}{c|}{VSE} & \multicolumn{4}{c|}{Ours} &
    VSE &
    \multicolumn{4}{|c}{Ours} \\
    \hline
    & \small{paper} & \small{current} & $\beta$=$1$ & $\beta$=$.75$ & $\beta = 0.5$ & $\beta = 0.25$ & \small{current} & $\beta$=$1$ & $\beta$=$.75$ & $\beta = 0.5$ & $\beta = 0.25$ \\
    \toprule
    
    \multicolumn{12}{c}{Sentence to image}\\
    \midrule
    
    r@1 & \underline{16.8} & 16.5 &
    23.0 {\greenbf{($+$6.2)}} &
    \textbf{24.9} {\greenbf{($+$8.1)}} &
    22.3 {\greenbf{($+$5.5)}} &
    21.3 {\greenbf{($+$4.5)}} &
    \underline{13.5} &
    \textbf{21.6} {\greenbf{($+$8.1)}} &
    20.3 {\greenbf{($+$6.8)}} &
    20.3 {\greenbf{($+$6.8)}} &
    19.5 {\greenbf{($+$6.0)}} \\
    
    r@5 & \underline{42.0} & 41.9 &
    49.3 {\greenbf{($+$7.3)}} &
    \textbf{52.3} {\greenbf{($+$10.3)}} & 
    48.3 {\greenbf{($+$6.3)}} &
    45.5 {\greenbf{($+$3.5)}} &
    \underline{36.6} &
    \textbf{48.8} {\greenbf{($+$12.2)}} &
    45.0 {\greenbf{($+$8.4)}} &
    43.7 {\greenbf{($+$7.1)}} &
    43.0 {\greenbf{($+$6.4)}} \\
    
    r@10 & \underline{56.5} &  54.4 &
    61.1 {\greenbf{($+$4.6)}} &
    \textbf{63.6} {\greenbf{($+$7.1)}} &
    58.4 {\greenbf{($+$1.9)}} &
    56.7 {\greenbf{($+$0.2)}} &
    \underline{49.0} &
    \textbf{59.5} {\greenbf{($+$10.5)}} &
    56.6 {\greenbf{($+$7.6)}} &
    55.4 {\greenbf{($+$6.4)}} &
    54.4 {\greenbf{($+$5.4)}} \\
    
    mrank & \underline{8} & 9 & 6 & \textbf{5} & 6 &
    7 &
    \underline{11} & \textbf{6} & 7 & 8 &
    8 \\
    \midrule
    
    \multicolumn{12}{c}{Image to sentence} \\
    \midrule
    
    r@1 & 23.0 & \underline{30.7} &
    \textbf{33.1} {\greenbf{($+$2.4)}} &
    30.7 {\greenbf{($+$0.0)}} &
    27.4 {\redbf{($-$3.3)}} &
    26.7 {\redbf{($-$4.0)}} &
    \underline{30.5} &
    \textbf{32.3} {\greenbf{($+$1.7)}} &
    24.9 {\redbf{($-$5.6)}} &
    23.0 {\redbf{($-$7.5)}} &
    21.8 {\redbf{($-$8.7)}} \\
    
    r@5 & 50.7 & \textbf{\underline{57.8}} &
    57.2 {\redbf{($-$0.6)}} &
    55.4 {\redbf{($-$2.4)}} &
    54.5 {\redbf{($-$3.3)}} &
    51.4 {\redbf{($-$6.4)}} &
    \underline{56.0} &
    \textbf{58.6} {\greenbf{($+$2.6)}} &
    52.3 {\redbf{($-$3.7)}} &
    48.4 {\redbf{($-$7.6)}} &
    49.8 {\redbf{($-$6.2)}} \\
    
    r@10 & 62.9 & \textbf{\underline{70.6}} &
    68.7 {\redbf{($-$1.9)}} &
    65.6 {\redbf{($-$5.0)}} &
    64.0 {\redbf{($-$6.6)}} &
    61.9 {\redbf{($-$8.7)}} &
    \textbf{\underline{68.9}} &
    68.1 {\redbf{($-$0.8)}} &
    63.6 {\redbf{($-$5.3)}} &
    62.8 {\redbf{($-$6.1)}} &
    61.3 {\redbf{($-$7.6)}} \\
    
    mrank & 5 & \textbf{\underline{4}} & \textbf{4} & \textbf{4} & \textbf{4} &
    5 &
    \textbf{\underline{4}} & \textbf{4} & 5 & 6 &
    6 \\

    \bottomrule
  \end{tabular}
  }
  \caption{Monolingual baseline (VSE) of~\newcite{Kirosetal2014b} and our MLMME model on the M30k$_\text{C}$ test set.
  Best monolingual results are underlined and best overall results appear in bold.
  We show improvements over the best monolingual baseline in parenthesis.
  Best viewed in colour.
  }
  \label{tbl:evaluation-multi30k-comparable-english-german}
\end{table*}

\section{Datasets}
\label{sec:data}

The original Flickr30k data set contains 30k images and
5 English sentence descriptions for each image~\cite{Youngetal2014}.
To train the NMT model in~\cref{sec:nmt} we use the \emph{translated Multi30k}, henceforth M30k$_\text{T}$, where for each of the 30k images in the original Flickr30k, one of its English descriptions is manually translated into German by a professional translator.
Training, validation and test sets contain 29k, 1014 and 1k images respectively, each accompanied by one translated sentence pair in English and German.
In all other experiments in~\cref{sec:ranking} we use the \emph{comparable Multi30k}, henceforth M30k$_\text{C}$, an expansion of the Flickr30k where 5 German descriptions were collected for each image in the original Flickr30k independently from the English descriptions~\cite{ElliottFrankSimaanSpecia2016}.
Training, validation and test sets contain 29k, 1014 and 1k images respectively, each accompanied by 5 English and 5 German sentences.
Since the M30k$_\text{C}$'s test set is not publicly available, we split its validation set in two as use the whole training set for training, the first 500 images and their corresponding bilingual sentences in the validation set for model selection and the remaining 514 images and bilingual sentences for evaluation.
Source and target languages were estimated over the entire vocabulary, i.e. 22,965 English and 34,036 German tokens.

\section{Experiments}
\label{sec:experiments}

For each language we train a separate 1024D encoder RNN with GRU.
Word embeddings are 620D and trained jointly with the model.
All non-recurrent matrices are initialised by sampling from a Gaussian, \mbox{$\mathcal{N}(0,0.01)$}, recurrent matrices are random orthogonal and bias vectors are all initialised to zero.
We apply dropout~\cite{Srivastavaetal2014} with a probability of $0.5$ in both text and image representations, which are in turn mapped onto a 2048D multi-modal embedding space.
We set the margin \mbox{$\alpha$ = $0.2$}.
Our models are trained using stochastic gradient descent with Adam~\cite{KingmaBa2015} with minibatches of 128 instances.

\section{Results}
\label{sec:results}

As our main baseline, we retrain~\newcite{Kirosetal2014b} monolingual models separately on English and German sentences (+images).

\paragraph{Image$\leftrightarrow$Sentence Ranking}
\label{sec:ranking}

In Table~\ref{tbl:evaluation-multi30k-comparable-english-german}, we show results for the monolingual VSE English and German models of~\newcite{Kirosetal2014b} and our MLMME models on the M30k$_\text{C}$ data set and evaluated on images and bilingual sentences.
Recall-at-$k$ (\emph{r@k}) measures the mean number of times the correct result appear in the top-$k$ retrieved entries and \emph{mrank} is the median rank.

First, we note that multilingual models show consistent improvements in ranking images given sentences.
All our models, regardless of the value of the hyperparameter $\beta$ ($=$ $.25, .5, .75, 1$) show strong improvements in recall@k (up to $+12.2$) and median rank---mrank in English reduced from 8 to 5 and in German from 11 to 6 in comparison to the best model by~\newcite{Kirosetal2014b}.
Nevertheless, when ranking sentences given images, results are unclear.
The best results achieved by our multilingual models, for both languages, are observed when \mbox{$\beta=1$}, with the recall@k deteriorating as we include more multilingual similarity, i.e. \mbox{$\beta$ = $.75, .5, .25$}, and the median rank also slightly increasing for English (from 4 to 5) and German (from 4 to 6).

\paragraph{Semantic Textual Similarity}
\label{sec:sts}

In the semantic textual similarity task, we use our model to compute the distance between a pair of sentences (distances are equivalent to cosine similarity and therefore lie in the $[0,1]$ interval).
Gold standard scores for all tasks are given in the $[0,5]$ interval, where $0$ means complete dissimilarity and $5$ complete similarity.
We simply use the cosine similarity distance and scale it by $5$, directly comparing it to the gold standard scores.
There is no SemEval data set including the German language, therefore we only use our English encoders to compute embedding vectors for both sentences in each test set.
We report results for the two in-domain similarity tasks in SemEval, specifically the image description similarity tasks from years 2014~\cite{Agirreetal2014} and 2015~\cite{Agirreetal2015}.

\begin{table}[t!]
  \centering
  \resizebox{0.9\linewidth}{!} {
  \begin{tabular}{ c|c|c|c|c|c|c }
    Test set & VSE & \multicolumn{4}{c|}{Our model} & SemEval \\
    & &$\beta$=$1$&$\beta$=$.75$&$\beta$=$.5$&$\beta$=$.25$& best \\
    \midrule
    
    \multicolumn{7}{c}{\textbf{in-domain data}}\\
    \midrule
    
    IMG$_\text{1}$ &
    .685 &
    .693 &
    .692 &
    \textbf{.727} &
    .717 &
    .821 \\
    \midrule
    
    IMG$_\text{2}$ &
    .687 &
    .752 &
    .728 &
    .776 &
    \textbf{.797} &
    .864 \\
    
    
    
    
    
    
    \bottomrule
    
  \end{tabular}
  }
  \caption{Pearson rank correlation scores for semantic textual similarities in
  two different SemEval test sets.
  IMG$_\text{1}$: image descriptions (2014),
  IMG$_\text{2}$: image descriptions (2015).
  Best overall scores (ours vs. monolingual baseline) in bold.
  }
  \label{tbl:evaluation-semeval}
\end{table}

In Table~\ref{tbl:evaluation-semeval}, we note that our MLMME model consistently improves on the monolingual baseline of~\newcite{Kirosetal2014b} in the two in-domain similarity tasks,
remaining competitive even compared to the best comparable SemEval model (differences $<$10\%).
We note that we only use the English side of our models in these two evaluations.
That means that the addition of the German encoder helps also the English encoder, specially for lower values of $\beta$ ($\beta \in \{.5,.25$\}).

\paragraph{Neural Machine Translation (NMT)}
\label{sec:nmt}

In this set of experiments, we train the attention-based model of~\newcite{BahdanauChoBengio2015}\footnote{\url{https://github.com/rsennrich/nematus}} on the M30k$_\text{T}$ training set to translate from English into German.
We use it to generate $n$-best lists ($n=20$) for each entry in the M30k$_\text{T}$ validation and test sets.
We use Kiros' monolingual VSE model trained on German sentences and images to compute the distance between them, and our MLMME models trained with $\beta \in \{.25,.5,.75,1\}$ to compute the distance between German and English sentences with $s_s(\cdot,\cdot)$, and between a German sentence and an image using $s_i(\cdot,\cdot)$, for all entries in the M30k$_\text{T}$ validation and test sets.
We then train an $n$-best list re-ranker on the M30k$_\text{T}$ validation set's $20$-best lists with k-best MIRA~\cite{CrammerSinger2003,CherryFoster2012}, and use the new distances as additional features to the original MT log-likelihood $p(Y \given X)$.
We finally apply the optimised weights to re-rank the test set's $20$-best lists.

\begin{table}[h!]
  \centering
  \resizebox{\linewidth}{!} {
    \begin{tabular}{rllll}
      & \textbf{Discriminative} & \textbf{BLEU} & \textbf{METEOR} & \textbf{TER} \\
      & \textbf{re-ranking?} &&& \\
      \toprule
      NMT & --- &
      30.9 &
      50.0 &
      50.7 \\
      
      + & VSE &
      35.2 {\green{(+4.3)}} &
      52.2 {\green{(+2.2)}} &
      44.8 {\green{(-5.9)}} \\
            
      + & MLMME, $\beta=1$ &
      35.2 {\green{(+4.3)}} &
      52.5 {\green{(+2.5)}} &
      44.8 {\green{(-5.9)}} \\
      
      + & MLMME, $\beta=0.75$ &
      35.2 {\green{(+4.3)}} &
      52.5 {\green{(+2.5)}} &
      44.8 {\green{(-5.9)}} \\
      
      + & MLMME, $\beta=0.5$ &
      35.0 {\green{(+4.1)}} &
      52.3 {\green{(+2.3)}} &
      44.9 {\green{(-5.8)}} \\
      
      + & MLMME, $\beta=0.25$ &
      \underline{\textbf{35.4}} {\greenbf{(+4.5)}} &
      \underline{\textbf{52.6}} {\greenbf{(+2.6)}} &
      \underline{\textbf{44.6}} {\greenbf{(-6.1)}} \\
      
      \bottomrule
    \end{tabular}
  }
  \caption{MT metrics computed for $1$-best translations generated with a baseline NMT model and for $20$-best lists generated by the same model, re-ranked using VSE and MLMME as discriminative features.}
  \label{tab:nmt_mlmm}
\end{table}

From Table~\ref{tab:nmt_mlmm}, incorporating features from discriminative models to re-rank $n$-best lists leads to strong improvements according to traditional MT metrics, e.g. 4.5 BLEU, 2.6 METEOR or 6.1 TER points~\cite{Papinenietal2002,DenkowskiLavie2014,Snoveretal2006}.
We note that the highest improvements were found for our MLMME model when $\beta=0.25$, i.e. using much of the multilingual similarities, even though all models show consistent gains in all three metrics.

\section{Conclusions}
\label{sec:conclusions}

We propose a new discriminative model that incorporates both multilingual and multi-modal similarities and introduce a modified noise-contrastive estimation function to optimise our model, which shows promising results in three different tasks.
Results obtained with the recently released Multi30k data set demonstrate that our model can learn meaningful multimodal embeddings, effectively making use of multilingual signals and leading to consistently better results in comparison to a comparable monolingual model.

In the future we will train our model on a many-languages setting, with images and descriptions in $\sim$10 languages.
Finally, we wish to study in more detail how these models can be exploited in NMT.


\bibliography{refs}
\bibliographystyle{acl_natbib}

\end{document}